%% file: main.tex
\documentclass[runningheads]{llncs}

\usepackage{eccv}

\usepackage{eccvabbrv}

\usepackage{graphicx}
\usepackage{booktabs}
\usepackage[ruled,vlined]{algorithm2e}
\usepackage[accsupp]{axessibility}  
\usepackage{multirow}
\usepackage{array}

\usepackage{hyperref}

\usepackage{orcidlink}
\begin{document}

\title{AMR-CCR: Anchored Modular Retrieval for Continual Chinese Character Recognition} 

\titlerunning{Abbreviated paper title}

\author{
Yuchuan Wu \and
Yinglian Zhu \and
Haiyang Yu \and
Ke Niu \and
Bin Li \textsuperscript{*} \and
Xiangyang Xue
}

\authorrunning{Y. Wu et al.}

\institute{
Fudan University, Shanghai, China \\
\email{\{ycwu24, ylzhu22, kniu22\}@m.fudan.edu.cn} \\
\email{\{hyyu20, libin, xyxue\}@fudan.edu.cn}
}

\maketitle

\begingroup
\renewcommand\thefootnote{}
\footnotetext{* Corresponding author.}
\endgroup

\begin{abstract}
Ancient Chinese character recognition is a core capability for cultural heritage digitization, yet real-world workflows are inherently non-stationary: newly excavated materials are continuously onboarded, bringing new classes in different scripts, and expanding the class space over time. We formalize this process as Continual Chinese Character Recognition (Continual CCR), a script-staged, class-incremental setting that couples two challenges: (i) scalable learning under continual class growth with subtle inter-class differences and scarce incremental data, and (ii) pronounced intra-class diversity caused by writing-style variations across writers and carrier conditions.
To overcome the limitations of conventional closed-set classification, we propose \textbf{AMR-CCR}, an anchored modular retrieval framework that performs recognition via embedding-based dictionary matching in a shared multimodal space, allowing new classes to be added by simply extending the dictionary. AMR-CCR further introduces a lightweight script-conditioned injection module (SIA+SAR) to calibrate newly onboarded scripts while preserving cross-stage embedding compatibility, and an image-derived multi-prototype dictionary that clusters within-class embeddings to better cover diverse style modes.
To support systematic evaluation, we build \textbf{EvoCON}, a six-stage benchmark for continual script onboarding, covering six scripts (OBC, BI, SS, SAC, WSC, CS), augmented with meaning/shape descriptions and an explicit zero-shot split for unseen characters without image exemplars.
\end{abstract}

\input{sec/1_introduction4}
\input{sec/2_related_work}
\input{sec/3_dataset_and_benchmark}
\input{sec/4_methodology}
\input{sec/5_experiments}

\input{sec/6_conclution}


%
%
\bibliographystyle{splncs04}
\bibliography{main}
\end{document}

%% file: sec/1_introduction4.tex
\section{Introduction}

\begin{figure*}
    \centering
    \includegraphics[width=1\linewidth]{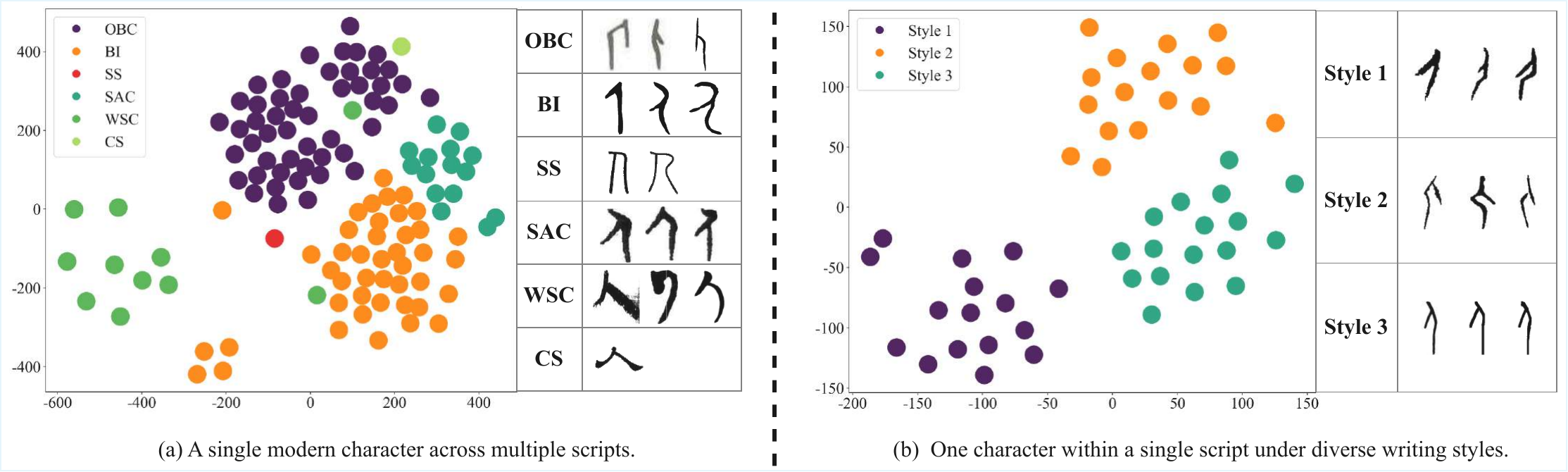}
    \caption{Embedding distribution and glyph examples for a modern character.}
    \label{introduction}
\end{figure*}

Ancient character recognition~\cite{gordin2024cured,fuentes2025recognition,swindall2021exploring} aims to automatically identify character forms from historical documents, rubbings of steles, and excavated materials, transforming massive collections of glyph images into structured resources that are searchable and analyzable. This capability serves as a fundamental building block for cultural heritage digitization and digital humanities research. In this paper, we focus on a highly representative branch—ancient Chinese Character Recognition (CCR)—where the task is distinguished by rich cross-script variations, long evolutionary trajectories, and a large character set.

Most existing studies on ancient CCR~\cite{Wang2026AttGraphDC,wang2025multi} adopt an offline closed-set classification setting: a classifier is trained on a fixed set of categories, assuming that training and test share the same label space. In contrast, real-world digitization workflows do not follow a one-shot closed-set construction. Newly excavated materials are continuously incorporated, introducing not only new scripts but also new character classes within existing scripts, causing the label space to expand over time~\cite{mejias2025technical}. To reflect this workflow, we formalize the problem as Continual Chinese Character Recognition (Continual CCR): data are onboarded in stages by script, and the class set is expanded in a class-incremental manner at each stage. Under this setting, continual onboarding brings two tightly coupled challenges (Fig.~\ref{introduction}): (i) the number of classes keeps growing, while inter-class differences can be extremely subtle, and the incremental process often comes with scarce samples; and (ii) a single class may exhibit diverse writing styles across different writers and carrier conditions, leading to a highly intra-class distribution.

In this setting, the conventional classification paradigm is inadequate. As the class space grows, the classifier head must be continually expanded and its decision boundaries re-calibrated: sequential fine-tuning can easily distort regions of old classes and cause forgetting, while full retraining becomes prohibitively expensive. Moreover, conventional classification enforces single-center class compactness, which conflicts with classes that exhibit diverse styles and yields fragile decision margins under continual updates.

To address the scalability challenge posed by a continually expanding class space, we move from closed-set classification to an embedding-based dictionary retrieval paradigm and propose AMR-CCR, an anchored modular retrieval framework. AMR-CCR represents characters in a shared multimodal embedding space and formulates recognition as similarity-based matching. Class knowledge is explicitly stored as dictionary entries, so new classes can be incorporated simply by adding them into the dictionary. Moreover, the multimodal embeddings can align visual features with textual features such as meaning and shape descriptions, enabling the model to leverage external knowledge for stronger discriminative power. We further introduce our designs to tackle the two challenges above:
\begin{itemize}
    \item
    To cope with distribution shift induced by continual class onboarding and the representation drift, we introduce a script-conditioned injection module (\textbf{SIA}+\textbf{SAR}) that performs lightweight, script-specific calibration for newly onboarded scripts while keeping the shared embedding space globally consistent. 

    \item
    To address the intra-class distribution, we construct an image-derived multi-prototype dictionary. 
    Concretely, we automatically cluster embeddings within each class and use multiple prototypes to cover distinct writing-style modes. 
\end{itemize}

To enable systematic and reproducible evaluation of Continual CCR, we introduce EvoCON, a six-stage benchmark extended from EVOBC~\cite{guan2024opendatasetevolutionoracle}. Following a continual-learning protocol, data are onboarded by script, covering Oracle Bone (OBC), Bronze Inscriptions (BI), Seal Script (SS), Spring-and-Autumn Characters (SAC), Warring States Characters (WSC), and Clerical Script (CS), with the stage order specified in the benchmark protocol. In addition, EvoCON includes an explicit zero-shot task to assess the model’s ability to handle unseen characters when no image exemplars are available.

Our main contributions are:
\begin{itemize}
    \item \textbf{Problem formulation.}
    Motivated by real-world digitization workflows, we formulate Ancient CCR as a script-staged, class-incremental problem. We further analyze the key challenges arising in this setting.

    \item \textbf{Framework.}
    We propose \textbf{AMR-CCR}, an anchored modular retrieval framework built on a shared multimodal embedding space.
    It maintains cross-stage embedding compatibility via script-conditioned injection (\textbf{SIA+SAR}), and models multiple styles in one class with an multi-prototype dictionary, enabling scalable continual onboarding and robust retrieval.

    \item \textbf{Benchmark.}
    We build \textbf{EvoCON}, a six-stage continual script-onboarding benchmark.
    EvoCON augments each sample with meaning/shape descriptions and provides an explicit zero-shot split, supporting systematic and reproducible evaluation of continual learning and zero-shot task.
\end{itemize}

%% file: sec/2_related_work.tex
\section{Related Work}

\subsection{Ancient Character Recognition}
Progress in ancient CCR has been driven by both new datasets and improved models~\cite{li2024comprehensivesurveyoraclecharacter,wang2024open, wang2022oracle, huang2019obc306}. EVOBC~\cite{guan2024opendatasetevolutionoracle} organizes oracle-bone character evolution across historical stages, making it possible to study extreme cross-script variation under a unified taxonomy; complementary resources further support recognition and decipherment, including collections of undeciphered oracle characters that emphasize open-world analysis beyond a fixed vocabulary~\cite{wang2024opendatasetoraclebone}. 
On the modeling side, most methods remain closed-set and classifier-centric under static training, typically improving robustness via stronger backbones and task-specific inductive biases such as component-aware modeling or noise-robust learning~\cite{Wang2026AttGraphDC,wang2025multi,wang2023gan}. A few works explore continual learning in this domain to accommodate newly arriving scripts/samples/categories~\cite{XU2024110283}, but they are usually studied within classifier pipelines and do not explicitly target dictionary-based recognition where long-term similarity structure and evidence retrieval are central. In contrast, our setting requires a unified protocol that supports continual onboarding under extreme shifts while preserving similarity relations for retrieval.

\subsection{Continual Learning for Recognition and Retrieval}
Classic continual learning is mainly studied in class-incremental recognition, with representative families including rehearsal, regularization/distillation, and exemplar-based methods~\cite{chaudhry2019tinyepisodicmemoriescontinual,Kirkpatrick_2017,rebuffi2017icarlincrementalclassifierrepresentation,buzzega2020darkexperiencegeneralcontinual}. While effective at stabilizing classifier outputs and preserving decision boundaries, such methods do not explicitly enforce the cross-stage similarity geometry required by dictionary-based recognition, where predictions rely on nearest neighbors and prototype compatibility.

A related line of research considers continual learning for retrieval embeddings, emphasizing representation drift as a key challenge: encoder updates may render previously stored prototypes or indexed features less compatible with the current embedding space, degrading nearest-neighbor retrieval. Representative directions include drift-aware compensation and feature-consistency regularization to maintain cross-stage compatibility~\cite{yu2020semanticdriftcompensationclassincremental,goswami2025query}. Different from these general-purpose formulations, we focus on a script-staged continual onboarding scenario with highly confusable near-duplicate classes and strong within-class glyph variation, and design a modular, anchor-based adaptation scheme tailored to preserving retrieval compatibility under such domain-specific shifts.

\subsection{Vision--Language Embeddings for Multimodal Retrieval}
Vision--language contrastive pretraining (e.g., CLIP~\cite{radford2021learning}, ALIGN~\cite{jia2021scaling}) learns shared image--text embeddings that support cross-modal retrieval and text-prototype zero-shot recognition. Such models provide a strong foundation for retrieval-style recognition, since they naturally output ranked candidates with interpretable visual/textual evidence and allow injecting priors via the choice of textual prototypes. Building on this line, Qwen3-VL~\cite{bai2025qwen3} and Qwen3-VL-Embedding~\cite{qwen3vlembedding} provide retrieval-oriented multimodal embeddings with a unified vector space for similarity search. 
However, most VLM embeddings are evaluated in static settings with fixed encoders and indexes. Although continual VLM training has only recently been explored~\cite{garg2023tic,zhu2023ctp}, maintaining dictionary compatibility under continual onboarding remains challenging: drift can invalidate previously stored prototypes or indexes and break similarity comparability across stages~\cite{goswami2025query}. This gap motivates designing drift-aware mechanisms that preserve cross-stage compatibility for dictionary-based matching in Continual CCR.

\begin{table}[t]
\centering
\caption{Statistics of splits across scripts. Img: number of images; Cls: number of classes; MCL: mean length of meaning captions; SCL: mean length of shape captions.}
\label{tab:dataset_statistic}
\setlength{\tabcolsep}{1.9pt}
\begin{tabular}{lcccccccccccc}
\toprule
& \multicolumn{4}{c}{Train} & \multicolumn{4}{c}{Test} & \multicolumn{4}{c}{Zero-shot} \\
\cmidrule(lr){2-5}\cmidrule(lr){6-9}\cmidrule(lr){10-13}
Script
& Img & Cls & MCL & SCL
& Img & Cls & MCL & SCL
& Img & Cls & MCL & SCL \\
\midrule
OBC & 67755 & 1632 & 284.0 & 47.2 & 7139 & 1555 & 298.4 & 46.8 & 413 & 104 & 314.8 & - \\
BI  & 40448 & 2427 & 289.0 & 46.9 & 1886 & 883  & 319.6 & 45.7 & 379 & 134 & 271.7 & - \\
SS  & 3881  & 3250 & 260.1 & 45.8 & 1561 & 1561 & 287.1 & 48.1 & 405 & 280 & 289.5 & - \\
SAC & 6961  & 1358 & 303.3 & 47.5 & 909  & 738  & 292.9 & 47.1 & 135 & 75  & 281.0 & - \\
WSC & 65136 & 3378 & 302.2 & 48.7 & 1239 & 937  & 287.1 & 48.1 & 442 & 180 & 324.3 & - \\
CS  & 1885  & 1772 & 331.5 & 46.6 & 299  & 299  & 366.6 & 46.3 & 353 & 280 & 282.7 & - \\
\midrule
Total & 186066 & 13817 & 295.0 & 47.1 & 13033 & 5973 & 308.6 & 47.0 & 2127 & 1053 & 294.0 & - \\
\bottomrule
\end{tabular}%
\end{table}

%% file: sec/3_dataset_and_benchmark.tex
\section{Dataset and Benchmark Protocol}

This section describes the dataset construction and benchmark protocol of \textbf{EvoCON}. Built on EVOBC~\cite{guan2024opendatasetevolutionoracle}, EvoCON is a multi-modal resource with images, script-aware labels, character-level meaning descriptions, and instance-level shape descriptions. We define two evaluation tracks: (i) a six-stage continual onboarding protocol for Continual CCR, evaluated with AA1--AA6 and FGT, and (ii) a strict zero-shot task evaluated with ZS.

\subsection{Dataset Construction}
\label{sec:dataset_construction}

Built on EVOBC (images + labels only) across six scripts (OBC, BI, SS, SAC, WSC, CS), EvoCON augments each sample with shape and meaning descriptions to address cross-script class confusability and support multimodal learning in a shared vision–text space.

\textbf{Description augmentation.}
Image-only labels provide limited supervision for Continual CCR when many scripts are involved.
To inject richer semantic (meaning) and structural (shape) priors beyond image-only signals, we augment EvoCON with two complementary Simplified Chinese textual descriptions:
(i) \emph{meaning descriptions}, built at the character level by extracting meanings from an open-source Chinese lexicon and then performing manual normalization to ensure consistency and coverage, and shared across scripts for the same modern character label; and
(ii) \emph{shape descriptions}, built at the image-instance level by prompting Qwen3-VL-Plus~\cite{bai2025qwen3} to describe discriminative glyph structures (e.g., layout, components, and stroke configurations) and subsequently applying human verification for correctness, capturing cross-script and intra-script variations~\cite{wang2024human}.
Further details on data sources, prompting templates, and quality control are provided in Suppl.\ Sec.~A.

\begin{figure*}
    \centering
   \includegraphics[width=1\linewidth]{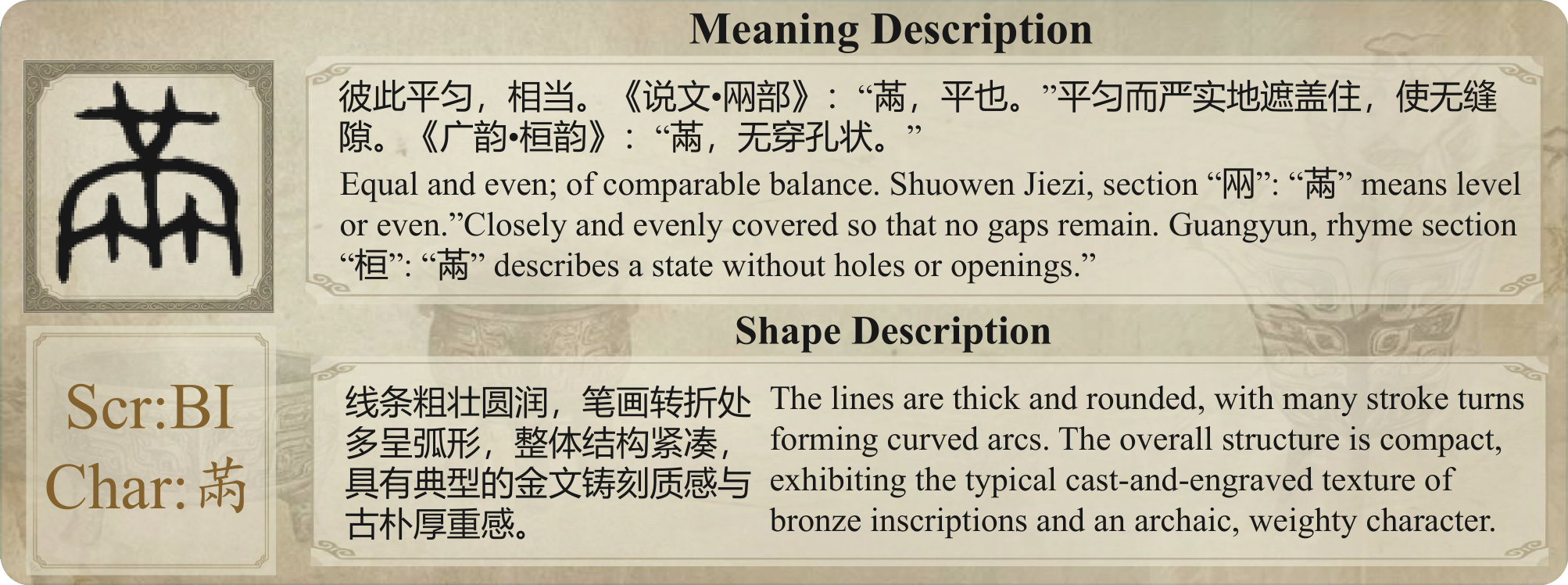}

   \caption{An EvoCON example with image, \texttt{scr} (script label), \texttt{char} (modern character label), and paired captions (meaning and shape descriptions).}
   \label{fig:dataset}
\end{figure*}

\textbf{Data partitioning for continual and zero-shot tracks.}
Since EVOBC does not provide an official train/test split, we construct script-wise splits for each stage.
For the continual onboarding track, we partition images within each script under a closed-set constraint, i.e., every class appearing in the test set also appears in the corresponding training set.
Due to the highly imbalanced data availability and class distributions across scripts, the resulting train/test ratio is not uniform (overall $\sim$14:1); detailed statistics are reported in Table~\ref{tab:dataset_statistic}.
For the zero-shot deciphering track, we additionally hold out low-resource characters (fewer than 5 images in total across the six scripts) entirely from training and evaluate only on their images.

\textbf{Dataset statistics.}
Table~\ref{tab:dataset_statistic} summarizes the data statistics across scripts for the continual onboarding and zero-shot tracks.
EvoCON is highly imbalanced in image volume: OBC, BI, and WSC contribute most training images, whereas SS and CS are much smaller, reflecting naturally skewed availability in cultural-heritage digitization.
Meanwhile, class coverage follows a different trend: some scripts (e.g., SS) have many classes but few images, producing a pronounced long tail and making continual onboarding harder due to highly confusable near-duplicate classes with limited per-class evidence.

EvoCON also provides two Simplified Chinese captions per sample.
Meaning captions are relatively long and stable across splits (MCL $\approx$ 260--330), serving as script-shared semantic priors, while shape captions are short and structure-oriented (SCL $\approx$ 46--49 in Train/Test), describing discriminative glyph patterns.
Figure~\ref{fig:dataset} illustrates examples; additional statistics are provided in Suppl.\ Sec.~B.

\subsection{Continual Onboarding Protocol}
\label{sec:protocol}

To simulate real-world deployment that requires unified recognition across multiple scripts arriving over time, we adopt a script-staged continual onboarding protocol in a class-incremental setting, where the label space expands stage by stage.
We use script-aware labels: a modern character is treated as a different class when it appears in different scripts.
This avoids collapsing systematic cross-script glyph transformations into within-class variation, and better matches the near-duplicate class confusability in continual onboarding.

\textbf{Stage ordering.}
We define a canonical 6-stage script-onboarding sequence for reproducibility:
CS $\rightarrow$ WSC $\rightarrow$ SAC $\rightarrow$ SS $\rightarrow$ BI $\rightarrow$ OBC.
This reverse-chronological order is chosen to better reflect practical digitization pipelines, where more standardized and easier-to-annotate scripts are typically integrated first, while earlier and more irregular scripts are incorporated later as data collection and expert annotations accumulate.

\textbf{Training stream.}
At stage $t$, the learner is trained \emph{only} on the training split of the current script $\mathcal{D}^{tr}_{t}$.
By default, data from previous stages is not accessible, while continual-learning methods may optionally retain a limited memory buffer of samples from earlier stages (following standard rehearsal protocols).
Access to any data from future stages is strictly prohibited during training.

\textbf{Evaluation schedule.}
After training stage $t$, we evaluate the model on the union of test splits from all observed stages, i.e., $\bigcup_{i=1}^{t}\mathcal{D}^{te}_{i}$.
This produces a stage-wise trajectory of performance that supports standard continual-learning analysis, including final accuracy and forgetting.



\subsection{Zero-shot Task}
\label{zeroshot task}
To extend the practical scope of CCR beyond closed-set classification of known classes to assisting the decipherment of previously unseen characters, we formulate a strict zero-shot task based on the zero-shot split defined in Section~\ref{sec:dataset_construction}.
At inference, the model receives an image-only query whose class is unseen during training (i.e., no image examples/prototypes of the target class are available).
The model is provided with auxiliary textual descriptions associated with candidate classes (e.g., meaning glosses), and is expected to leverage both visual evidence and this side information to identify the correct class.
We evaluate by ranking candidate classes and report ZS@1 and ZS@20 (Top-1/Top-20).


\begin{figure*}[t]
    \centering
   \includegraphics[width=1\linewidth]{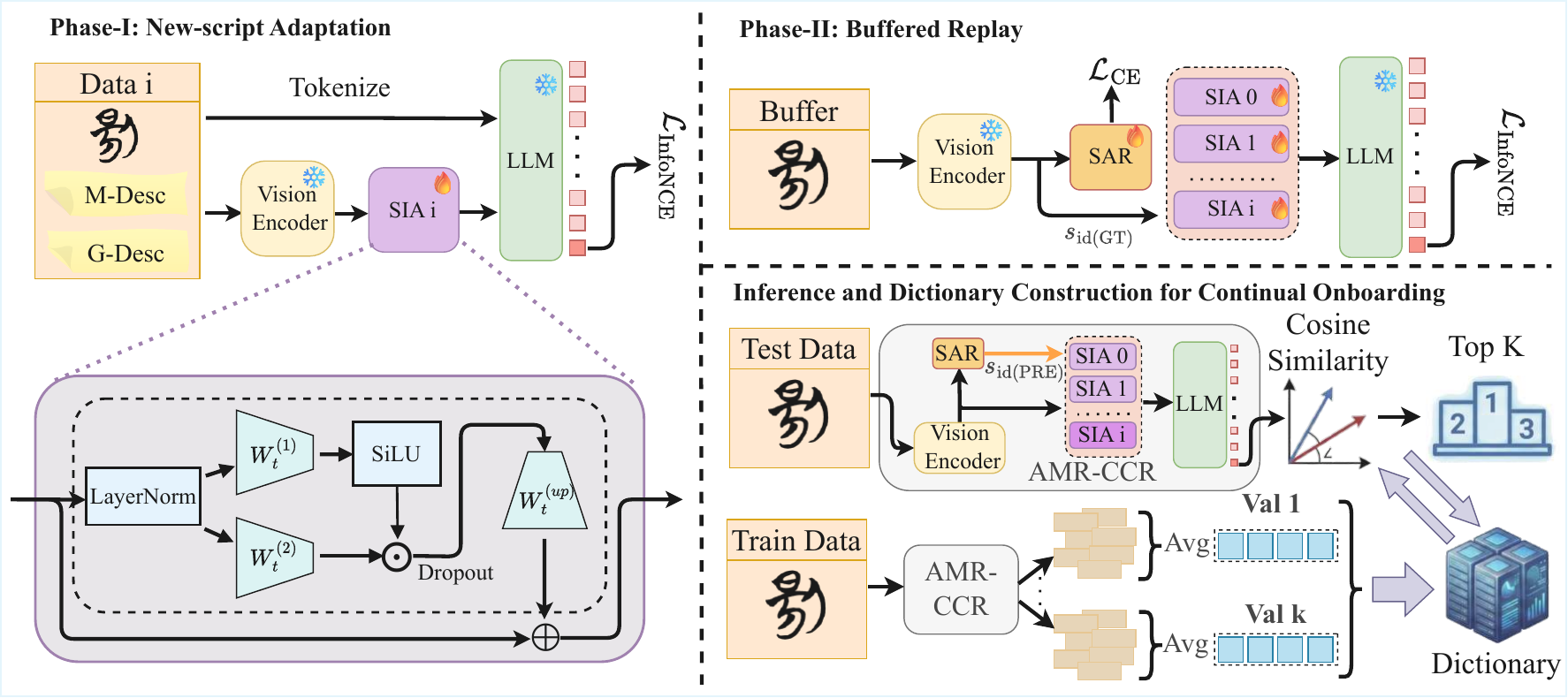}

   \caption{\textbf{Overview of AMR-CCR.}}
   \label{architecture}
\end{figure*}

%% file: sec/4_methodology.tex
\section{Methodology}
\label{sec:method}

We address Continual CCR using a retrieval-centric embedding framework (Fig.~\ref{architecture}). Our method, AMR-CCR, performs recognition via dictionary-based matching in a shared vision--language embedding space, returning ranked candidates with visual/textual evidence and supporting an expandable label space. To remain reliable under continual onboarding with highly confusable near-duplicate classes, AMR-CCR preserves cross-stage embedding compatibility by freezing a strong vision--language backbone as a stability anchor and adapting through lightweight script-conditioned modules(SIA and SAR). It further handles high within-class writing variation with an image-derived multi-prototype dictionary, and enables strict zero-shot deciphering by retrieving against a text dictionary of meaning descriptions when no image prototypes exist.

\subsection{Training Data Preparation}
\label{sec:data_preparation}

At continual stage $t$, we prepare two training sets for the two training phases: a new-script set for Phase-I and a buffered replay set for Phase-II.

\textbf{Phase-I (new script set).}
For the newly onboarded script at stage $t$, we generate multiple training pairs per anchor image by repeatedly sampling a single positive for each training instance.
Specifically, each image serves as an anchor, and for every instance we sample one positive from three sources:
(i) another image from the same class,
(ii) the meaning description, or
(iii) the shape description associated.
Importantly, each instance uses exactly one source (no concatenation),
we draw positives with a ratio of $8{:}1{:}1$ for visual, meaning, and shape sources, respectively, and place the resulting training pairs into the same mini-batches for joint image--image and image--text training.

\textbf{Phase-II (buffered replay set).}
We maintain a fixed-size memory buffer of 10{,}000 images, updated after each stage to store an approximately script-balanced subset from all observed scripts up to stage $t$ (images only).
Each buffered image retains its script $s_{\mathrm{id}}$ for training-time supervision, while $s_{\mathrm{id}}$ is not assumed at inference.
During replay, each buffered image serves as an anchor and a positive image is sampled from the same script-aware class within the buffer to form contrastive training pairs.

\subsection{Model Architecture}
\label{sec:method_arch}

We adopt Qwen3-VL-Embedding~\cite{qwen3vlembedding} as the shared vision--language embedding backbone, which provides a strong unified image--text embedding space for multimodal retrieval.
On top of this frozen backbone, we build lightweight script-conditioned modules (SIA and SAR) to support script-specific calibration while maintaining a shared embedding space across stages (Fig.~\ref{architecture}).

\noindent\textbf{Script-Interface Adapter (SIA).}
Given a frozen visual embedding $\mathbf{e}\in\mathbb{R}^{D}$, SIA extracts script-specific cues and performs a lightweight, script-conditioned calibration while preserving the shared embedding space.

\smallskip
To ensure stable behavior across scripts, the embedding is first normalized:
\begin{equation}
\mathbf{h}=\mathrm{LN}(\mathbf{e}).
\end{equation}

\smallskip
A pair of script-conditioned low-rank projections ($r\ll D$) then produces compact offset candidates:
\begin{equation}
\mathbf{a}=W^{(1)}_t\mathbf{h},\qquad \mathbf{b}=W^{(2)}_t\mathbf{h},
\qquad W^{(1)}_t,W^{(2)}_t:\mathbb{R}^{D}\rightarrow\mathbb{R}^{r}.
\end{equation}

\smallskip
The two offsets are fused with a SwiGLU-style gate, lifted back to $\mathbb{R}^{D}$, and applied as a scaled residual correction:
\begin{equation}
\mathbf{z}=\sigma(\mathbf{a})\odot \mathbf{b}\in\mathbb{R}^{r},
\end{equation}
\begin{equation}
\Delta\mathbf{e}=W^{\mathrm{up}}_t\mathbf{z},\qquad 
\mathbf{e}'=\mathbf{e}+\alpha_t\Delta\mathbf{e},
\qquad W^{\mathrm{up}}_t:\mathbb{R}^{r}\rightarrow\mathbb{R}^{D}.
\end{equation}

\noindent\textbf{Script-Aware Routing (SAR).}
At inference time, the script identity is unknown, while SIAs are script-specific. SAR therefore introduces a lightweight MLP router that selects the most suitable adapter directly from the frozen visual embedding. We adopt hard routing so that each query activates exactly one SIA, avoiding the feature dilution that may arise from softly mixing multiple adapters and keeping inference cost constant as the number of scripts grows. Note that SAR does not predict character classes; it only chooses a script-conditioned calibration prior to dictionary retrieval, and the final recognition is still determined by similarity matching in the shared embedding space.

\smallskip
\emph{(1) Router prediction.}
Given a frozen embedding $\mathbf{e}$, the router $g(\cdot)$ produces a distribution over the adapters available up to stage $t$:
\begin{equation}
\mathbf{p}=g(\mathbf{e})\in\mathbb{R}^{t}, \qquad \sum_{j=1}^{t}\mathbf{p}_j=1.
\end{equation}
The router is trained with script labels using cross-entropy, encouraging it to capture script cues that help disambiguate near-duplicate classes.

\smallskip
\emph{(2) Hard routing.}
At inference, a single adapter is selected as
\begin{equation}
k=\arg\max_{j}\mathbf{p}_j,
\end{equation}
and only $\mathrm{SIA}_k$ is applied to obtain the adapted embedding for retrieval.

\subsection{Training Recipe}
\label{sec:method_training}

We optimize AMR-CCR stage by stage.
Throughout training, the Qwen3-VL-Embedding backbone is kept frozen as the \emph{Stability Anchor (SA)} to preserve a shared embedding space across stages.
At each stage $t$, we use a two-phase training procedure with complementary goals:
(i) \emph{new-script adaptation}, where we train the stage-$t$ SIA using only the current script to capture script-specific calibration offsets without drifting the shared space; and
(ii) \emph{buffered replay}, where we train on a script-balanced memory buffer (images only) to interleave previously seen scripts, thereby maintaining cross-stage compatibility and reducing forgetting (and providing supervision for SAR when applicable).


\textbf{Phase-I: new-script adaptation.}
At stage $t$, we freeze the Stability Anchor (SA), the router $g(\cdot)$, and all previously learned adapters $\{\mathrm{SIA}_1,\ldots,\mathrm{SIA}_{t-1}\}$, and optimize only the newly introduced $\mathrm{SIA}_t$ using the Phase-I data in Sec.~\ref{sec:data_preparation}.
For each anchor image $x_i$, the final retrieval embedding is computed by inserting the adapter at the interface of the frozen backbone:
\begin{equation}
\hat{\mathbf{e}}_i
=
E_{\mathrm{emb}}\!\Big(
\mathrm{SIA}_t\big(E_{\mathrm{vis}}(x_i)\big)
\Big),
\end{equation}
where $E_{\mathrm{emb}}(\cdot)$ denotes all frozen modules after the SIA insertion point in Qwen3-VL-Embedding.
We contrast $\hat{\mathbf{e}}_i$ against a sampled positive embedding $\mathbf{u}_i$ under a unified InfoNCE objective with in-batch negatives:
\begin{equation}
\mathcal{L}^{(I)}_{\mathrm{InfoNCE}}
= -\log
\frac{\exp(\mathrm{sim}(\hat{\mathbf{e}}_i,\mathbf{u}_i)/\tau)}
{\sum_{j=1}^{B}\exp(\mathrm{sim}(\hat{\mathbf{e}}_i,\mathbf{u}_j)/\tau)}.
\end{equation}
Here $\mathrm{sim}(\cdot,\cdot)$ is cosine similarity on $\ell_2$-normalized embeddings, $B$ is the number of candidates in the mini-batch, and $\tau$ is a temperature.
The positive $\mathbf{u}_i$ is the embedding of exactly one sampled positive (Sec.~\ref{sec:data_preparation}).
All other candidate embeddings $\{\mathbf{u}_j\}$ in the mini-batch (images and texts) act as in-batch negatives for anchor $i$ except the matched positive.

\textbf{Phase-II: buffered replay.}
After Phase-I, we perform buffered replay using \emph{only} the memory buffer (10{,}000 images; Sec.~\ref{sec:data_preparation}) to (i) align the SIA bank across scripts and mitigate forgetting, and (ii) train the Script-Aware Routing (SAR) module for adapter selection at inference.

\smallskip
\noindent\emph{(a) Replay loss for SIA alignment.}
For each buffered pair $(x_i, x_i^+)$ from the same script-aware class, we compute their embeddings using the SIA corresponding to the image script $s_{\mathrm{id}}$ and minimize an image--image InfoNCE loss with in-batch negatives:
\begin{equation}
\mathcal{L}^{(buf)}_{\mathrm{InfoNCE}}
= -\log
\frac{\exp(\mathrm{sim}(\hat{\mathbf{e}}_i,\hat{\mathbf{e}}^{+}_i)/\tau)}
{\sum_{j=1}^{B}\exp(\mathrm{sim}(\hat{\mathbf{e}}_i,\hat{\mathbf{e}}_j)/\tau)}.
\end{equation}

\smallskip
\noindent\emph{(b) Routing loss for SAR.}
In parallel, we train the router $g(\cdot)$ to predict the SIA index from the frozen visual embedding $\mathbf{v}=E_{\mathrm{vis}}(x)$:
\begin{equation}
\mathbf{p}=g(\mathbf{v})\in\mathbb{R}^{t},
\qquad
\mathcal{L}_{\mathrm{CE}}=-\log \mathbf{p}_{s_{\mathrm{id}}}.
\end{equation}

\smallskip
During Phase-II, $\mathcal{L}^{(buf)}_{\mathrm{InfoNCE}}$ updates only the SIA bank $\{\mathrm{SIA}_1,\ldots,\mathrm{SIA}_t\}$, while $\mathcal{L}_{\mathrm{CE}}$ updates only the router $g(\cdot)$; the stability anchor (SA) remains frozen.
At inference, ground-truth script identifiers are unavailable and the SIA is selected by SAR (Sec.~\ref{sec:method_arch}).

\begin{algorithm}[t]
\caption{Auto-$K$ Multi-Prototype Dictionary Construction}
\label{alg:autoK_short}
\KwIn{Train embeddings $\{(\mathbf{e}_i,y_i)\}_{i=1}^N$ with \emph{script-aware} labels $y_i$ (embeddings computed via SAR routing), $K_{\max}{=}32$, sample cap $S{=}256$}
\KwOut{Prototype bank $\mathbf{P}$ (flattened), pointer array $\boldsymbol{\pi}$}

Group embeddings by class: $\mathbf{E}_y \leftarrow \{\mathbf{e}_i: y_i{=}y\}$.\;
Initialize $\mathbf{P}\leftarrow[\ ],\ \boldsymbol{\pi}\leftarrow[0]$\;

\ForEach{class $y$}{
  $K_y^{\max}\leftarrow \min(K_{\max},\lfloor\sqrt{|\mathbf{E}_y|}\rfloor)$\;
  
  \eIf{$K_y^{\max}<2$}{
    $K_y\leftarrow 1$\;
  }{
    Sample $\tilde{\mathbf{E}}_y\subseteq \mathbf{E}_y$ with $|\tilde{\mathbf{E}}_y|=\min(|\mathbf{E}_y|,S)$\;
    $K_y \leftarrow \arg\max_{k\in[2,K_y^{\max}]} 
    \mathrm{Silhouette}_{\cos}(\tilde{\mathbf{E}}_y,\mathrm{SphKMeans}(\tilde{\mathbf{E}}_y,k))$\;
  }

  $\mathbf{C}_y \leftarrow \mathrm{SphKMeans}(\mathbf{E}_y, K_y)$\;
  Append $\mathbf{C}_y$ to $\mathbf{P}$; append new prefix sum to $\boldsymbol{\pi}$\;
}

\end{algorithm}



  



\subsection{Inference}
\label{sec:method_infer}

We adopt a unified retrieval-based inference protocol.
Given an image-only query, the model first computes a script-conditioned embedding via SAR and SIA, and then ranks candidate classes by cosine similarity against a pre-built dictionary.
The two tasks differ only in dictionary construction: image prototypes for continual onboarding and text embeddings of meaning descriptions for zero-shot deciphering.

\textbf{Inference protocol.}
Given a query image $x$, we compute its script-conditioned embedding by routing to the most suitable SIA:
\begin{equation}
k=\arg\max_j g(E_{\mathrm{vis}}(x))_j,\qquad
\mathbf{q}=\mathrm{norm}\!\left(
E_{\mathrm{emb}}\!\Big(\mathrm{SIA}_{k}\big(E_{\mathrm{vis}}(x)\big)\Big)
\right).
\end{equation}

\noindent Recognition is performed by cosine similarity between the query embedding and dictionary prototypes, and classes are ranked to produce Top-$K$ predictions.

\textbf{Dictionary for continual onboarding.}
For the continual track, we build an expandable multi-prototype dictionary from training embeddings of all observed scripts, representing each script-aware class $y$ with $K_y$ spherical $k$-means prototypes (Auto-$K$) to capture within-class variation.
Embeddings are computed using the same SAR-based routing as inference (i.e., selecting the predicted SIA).
Algorithm~\ref{alg:autoK_short} provides the full construction and retrieval details.

\textbf{Dictionary for zero-shot task.}
For the zero-shot track, the dictionary contains only text embeddings of candidate meaning descriptions.
Since meaning texts provide no script identity, we first use SAR to predict the query script and select the corresponding SIA for image embedding extraction.
The resulting image embedding is then matched against the meaning-text dictionary by cosine similarity to produce ZS@1 and ZS@20 results.

%% file: sec/5_experiments.tex
\section{Experiments}

\subsection{Datasets and Metrics}
All experiments are conducted on EvoCON following the benchmark protocol.
We report two tracks. Continual onboarding (6 stages): after each stage $t$, we evaluate on all observed scripts and summarize Top-1/Top-10 performance with stage-wise average accuracy (AA1--AA6) and final-stage forgetting (FGT).
Zero-shot task: we rank unseen-class queries against a meaning-text dictionary and report ZS@1 and ZS@20.
Formal metric definitions are provided in Suppl.\ Sec.~C.

\begin{table}[t]
\centering
\caption{Results under the continual onboarding protocol. Seq FT and Joint FT denote fine-tuning performed sequentially over six tasks and jointly on data from all six tasks, respectively. Pretraining and joint training yield a single model weight set, hence only AA6 performance is shown.}
\label{tab:continual_singlecol_pct}
\setlength{\tabcolsep}{2.2pt}
\renewcommand{\arraystretch}{1.12}
\resizebox{\columnwidth}{!}{%
\begin{tabular}{lccccccc}
\toprule
Method & AA1\,{$\uparrow$} & AA2\,{$\uparrow$} & AA3\,{$\uparrow$} & AA4\,{$\uparrow$} & AA5\,{$\uparrow$} & AA6\,{$\uparrow$} & FGT\,{$\downarrow$} \\
\midrule
\multicolumn{8}{c}{\textit{Resnet-50~\cite{he2016deep} Based}} \\
\cmidrule(lr){2-8}
Seq FT & 0.00/3.34 & 8.64/20.37 & 1.98/57.57 & 0.73/1.40 & 12.17/13.04 & 14.69/16.34 & 16.81/27.89 \\
Joint FT   & -- & -- & -- & -- & -- & 42.38/62.67 & -- \\

\midrule
\multicolumn{8}{c}{\textit{Qwen3-VL-Embedding-8B~\cite{qwen3vlembedding} Based}} \\
\cmidrule(lr){2-8}
Frozen      & -- & -- & -- & -- & -- & 36.68/55.83 & -- \\
Joint FT  & -- & -- & -- & -- & -- & 41.14/66.93 & -- \\
Seq FT   & 67.22/87.62 & 24.13/43.73 & 21.63/48.34 & 22.91/49.06 & 16.53/38.10 & 28.68/55.25 & 9.52/9.70 \\

ER~\cite{wu2019memoryreplayganslearning}         & 66.88/86.62 & 36.01/54.45 & 29.40/50.55 & 27.74/52.40 & 26.29/49.51 & 34.74/59.09 & 3.09/4.44 \\
LwF~\cite{li2017learningforgetting}       & 68.18/88.20 & 38.50/56.95 & 31.04/53.17 & 30.32/54.98 & 28.78/52.00 & 36.82/61.13 & 2.81/3.65 \\
EWC~\cite{Kirkpatrick_2017}        & \underline{68.43}/\underline{88.79} & 38.85/56.85 & 32.03/52.93 & 30.65/55.15 & 29.10/52.28 & 37.26/61.49 & 2.94/3.71 \\
iCaRL~\cite{rebuffi2017icarlincrementalclassifierrepresentation}      & 66.22/86.28 & 36.46/52.24 & 30.64/49.69 & 28.35/51.26 & 25.84/49.10 & 31.74/56.15 & 2.97/3.34 \\
DER++~\cite{buzzega2020darkexperiencegeneralcontinual} & \textbf{68.67}/88.43 & \underline{40.40}/64.11 & 35.51/59.13 & 40.80/64.39 & 42.05/64.80 & 45.85/68.01 & 2.54/3.15 \\

\midrule
Ours (2B)  & 57.86/84.95 & 39.52/\underline{65.97} & \underline{37.61}/\underline{63.56} & \underline{46.12}/\underline{70.62} & \underline{46.87}/\underline{70.37} & \underline{50.43}/\underline{72.86} & \textbf{1.16}/\textbf{1.70} \\
Ours (8B)  & 68.40/\textbf{90.02} & \textbf{55.49}/\textbf{84.60} & \textbf{50.67}/\textbf{80.29} & \textbf{56.12}/\textbf{83.61} & \textbf{57.06}/\textbf{81.82} & \textbf{58.59}/\textbf{83.09} & \underline{1.85}/\underline{2.66} \\

\bottomrule
\end{tabular}%
}
\end{table}

\subsection{Implementation Details}
\label{sec:impl}
All methods are implemented in PyTorch with mixed precision, using Qwen3-VL-Embedding-2B and 8B~\cite{qwen3vlembedding} as the frozen backbone.
We set the SIA bottleneck rank to $r{=}64$ and initialize the residual scale as $\alpha{=}0.5$; the replay memory stores 10{,}000 images (images only) with approximately balanced script sampling.
Unless otherwise stated, we train with AdamW (learning rate $1\mathrm{e}{-}4$, weight decay $0.1$) using a cosine learning-rate scheduler with a warmup ratio of $0.01$, batch size $128$, for $6$/$5$ epochs (Phase-I/II); InfoNCE uses in-batch negatives with temperature $\tau{=}0.1$.
For dictionary-based inference, we use $\ell_2$ normalization and cosine retrieval with an auto-$K$ multi-prototype dictionary, setting $K_{\max}{=}32$ and per-class sample cap $S{=}256$.
We focus on widely used continual-learning baselines; many recent methods do not align with our pipeline, making faithful reproduction and fair comparison non-trivial, so we leave them for future work.
Details for reproducing other baselines are provided in Suppl.\ Sec.~D.

\subsection{Benchmark Results}
As the zero-shot task is not applicable to the baselines, we report zero-shot results in Section~\ref{ablation}. Qualitative retrieval examples and failure-cases are provided in Suppl.\ Sec.~E.
Table \ref{tab:continual_singlecol_pct} shows that under the six-stage continual onboarding setting of Continual CCR, the closed-set classification paradigm is prone to instability and severe forgetting. With ResNet-50, sequential fine-tuning reaches only 14.69/16.34 on AA6 with severe forgetting, while joint training improves to 42.38/62.67. As an oracle upper bound, this gap indicates that the backbone alone is insufficient for Continual CCR.

On Qwen3-VL-Embedding-8B, Joint FT consistently outperforms the Frozen model on AA6, indicating that the pretrained embeddings are fairly robust but still benefit from task-specific adaptation. However, naive sequential fine-tuning (Seq FT) drops to 28.68/55.25 at AA6, with FGT as high as 9.52/9.70, showing that continual updates can break cross-stage embedding compatibility. After incorporating continual learning strategies such as ER/LwF/EWC/iCaRL/DER++, forgetting is significantly reduced to around 2.5–3.1 (Top-1). DER++ performs best among them (AA6 45.85/68.01, FGT 2.54/3.15), yet its accuracy remains limited, indicating that the bottleneck is not only forgetting but also fine-grained confusion caused by dense near-duplicate classes and intra-class multi-modality.

In contrast, AMR-CCR achieves the best trade-off between accuracy and stability. Ours (8B) reaches 58.59/83.09 at AA6, outperforming DER++ by +12.74/+15.08, while keeping forgetting low at 1.85/2.66. This gain stems from two components: SIA+SAR preserves cross-stage embedding compatibility while enabling script-specific calibration, preventing new scripts from disrupting previously learned similarity structures; and the multi-prototype dictionary covers diverse writing styles and media within each class, mitigating the averaging bias of a single mean prototype, thereby substantially reducing confusion among near-duplicate classes and improving the quality of Top-1/Top-10 retrieval rankings.

\begin{table}[t]
\centering
\caption{Ablation results under the continual protocol and the zero-shot (ZS) task. All values are percentages ($\times 100$) with one decimal. Gold denotes using ground-truth routing, RS denotes random sampling.}
\label{tab:ablation_singlecol_pct}
\setlength{\tabcolsep}{2.0pt}
\renewcommand{\arraystretch}{1.10}
\resizebox{\columnwidth}{!}{%
\begin{tabular}{>{\raggedright\arraybackslash}p{1.55cm} l cccccccc}
\toprule
 & Variant & AA1\,{$\uparrow$} & AA2\,{$\uparrow$} & AA3\,{$\uparrow$} & AA4\,{$\uparrow$} & AA5\,{$\uparrow$} & AA6\,{$\uparrow$} & FGT\,{$\downarrow$} & ZS\,{$\uparrow$} \\
\midrule

\multirow[t]{1}{=}{\textit{MS}}
& 2B & 57.9/85.0 & 39.5/66.0 & 37.6/63.6 & 46.1/70.6 & 46.9/70.4 & 50.4/72.9 & 1.2/1.7 & 20.9/38.6 \\
\midrule

\multirow[t]{2}{*}{\parbox[t]{1.55cm}{\vspace{0pt}\textit{Modules}}}
& w/o SIA \& SAR & 67.2/87.6 & 24.1/43.7 & 21.6/48.3 & 22.9/49.1 & 16.5/38.1 & 28.7/55.3 & 9.5/9.7 & --/-- \\
& w/o SAR (Gold) & 68.9/90.3 & 56.2/85.7 & 52.4/83.4 & 57.5/85.8 & 59.3/85.3 & 61.2/86.8 & 0.9/1.7 & 15.9/48.8 \\
\midrule

\multirow[t]{2}{*}{\parbox[t]{1.55cm}{\vspace{0pt}\textit{BufS}}}
& 5k  & 68.2/80.8 & 55.1/84.2 & 49.2/79.9 & 54.3/81.5 & 55.1/82.8 & 57.2/80.8 & 2.8/3.6 & 14.3/43.9 \\
& 15k & 68.4/90.1 & 55.7/84.9 & 50.7/80.4 & 56.3/83.9 & 57.9/82.2 & 59.1/83.8 & 1.7/2.4 & 15.2/47.0 \\
\midrule

\multirow[t]{2}{*}{\parbox[t]{1.55cm}{\vspace{0pt}\textit{PC}}}
& Mean & 67.6/87.9 & 41.0/58.1 & 40.9/61.9 & 48.8/69.0 & 47.1/68.1 & 48.5/70.2 & 1.6/2.2 & --/-- \\
& RS(M=8) & 68.2/88.0 & 42.2/59.1 & 40.6/61.8 & 48.5/69.0 & 45.1/63.2 & 46.4/65.4 & 1.6/2.1 & --/-- \\
\midrule

\multirow[t]{1}{=}{\textit{Data-S1}}
& Image-only & 20.7/46.5 & 16.9/40.0 & 25.7/49.7 & 35.5/59.5 & 37.1/61.1 & 43.4/65.8 & 1.1/1.9 & 9.9/25.9 \\
\midrule

& Ours (8B) & 68.4/90.0 & 55.5/84.6 & 50.7/80.3 & 56.1/83.6 & 57.1/81.8 & 58.6/83.1 & 1.9/2.7 & 15.2/46.8 \\
\bottomrule
\end{tabular}
}
\end{table}

\subsection{Ablation Results}
\label{ablation}
Table~\ref{tab:ablation_singlecol_pct} validates the effectiveness of AMR-CCR from five aspects—model size, key components, buffer size, prototype construction, and training data—while evaluating both the continual protocol (AA/FGT) and the zero-shot task (ZS).

\noindent\textbf{(1) Model size (MS).}
The 2B model achieves 50.4/72.9 at AA6, substantially lower than the 8B model (58.6/83.1), indicating that larger capacity is more critical for fine-grained discrimination among dense near-duplicate classes. Meanwhile, the 2B model exhibits lower forgetting (FGT 1.2/1.7 vs.\ 1.9/2.7), reflecting a typical capacity--stability trade-off. ZS shows a similar tendency (2B: 20.9/38.6; 8B: 15.2/46.8), suggesting complementary strengths between Top-1 accuracy and Top-20 recall.

\noindent\textbf{(2) Modules.}
Removing both SIA and SAR drops AA6 to 28.7/55.3 and increases forgetting to 9.5/9.7, consistent with naive sequential fine-tuning. Therefore, script-conditioned injection and routing are required to preserve cross-stage embedding compatibility. Gold script routing (w/o SAR (Gold)) achieves AA6 61.2/86.8, forgetting 0.9/1.7, and ZS 15.9/48.8, which we treat as an oracle upper bound. The full model is close to this bound, implying that the learned router attains near-oracle script selection in this setting.

\noindent\textbf{(3) Buffer size (BufS).}
Increasing the buffer from 5k to 15k raises AA6 from 57.2/80.8 to 59.1/83.8, while lowering forgetting from 2.8/3.6 to 1.7/2.4. ZS also improves from 14.3/43.9 to 15.2/47.0, suggesting that stronger cross-stage replay better stabilizes the similarity structure and reduces boundary drift caused by near-duplicate classes. Notably, the gain from 10k to 15k is marginal (+0.5/+0.7), whereas reducing the buffer to 5k leads to a clear drop (-1.4/-2.3), supporting 10k as a good trade-off between performance and memory cost.

\noindent\textbf{(4) Prototype construction (PC).}
Using a single mean prototype (\emph{Mean}) or random sampling (\emph{RS}, $M{=}8$) substantially degrades AA6 (48.5/70.2 and 46.4/65.4, respectively), while forgetting remains similar ($\sim$1.6/2.1). This suggests that the primary bottleneck lies in insufficient coverage of diverse writing styles and resulting confusion with neighboring (near-duplicate) classes, rather than forgetting alone—supporting the importance of more structured and representative prototype construction for retrieval-based recognition.

\noindent\textbf{(5) Training data (Data-S1).}
Without textual supervision in Phase-I, AA6 drops to 43.4/65.8 and ZS to 9.9/25.9, demonstrating that vision--language alignment with meaning/shape descriptions not only improves standard retrieval discrimination but also directly determines text-only dictionary matching in the zero-shot setting.

%% file: sec/6_conclution.tex
\section{Conclusion}
We formalized ancient Chinese character recognition in real-world digitization as Continual CCR, a script-staged, class-incremental setting with continual class growth and strong style diversity. To move beyond closed-set classification, we proposed \textbf{AMR-CCR}, a multimodal retrieval framework that performs dictionary-based matching, supports plug-in class addition, and maintains cross-stage compatibility via script-conditioned injection (SIA+SAR) and an image-derived multi-prototype dictionary. We further introduced \textbf{EvoCON}, a six-stage benchmark with meaning/shape descriptions and a zero-shot split for systematic evaluation.

%% file: main.bib
@misc{guan2024opendatasetevolutionoracle,
      title={An open dataset for the evolution of oracle bone characters: EVOBC}, 
      author={Haisu Guan and Jinpeng Wan and Yuliang Liu and Pengjie Wang and Kaile Zhang and Zhebin Kuang and Xinyu Wang and Xiang Bai and Lianwen Jin},
      year={2024},
      eprint={2401.12467},
      archivePrefix={arXiv},
      primaryClass={cs.AI},
      url={https://arxiv.org/abs/2401.12467}, 
}

@article{XU2024110283,
title = {Large-scale continual learning for ancient Chinese character recognition},
journal = {Pattern Recognition},
volume = {150},
pages = {110283},
year = {2024},
issn = {0031-3203},
doi = {https://doi.org/10.1016/j.patcog.2024.110283},
url = {https://www.sciencedirect.com/science/article/pii/S0031320324000347},
author = {Yue Xu and Xu-Yao Zhang and Zhaoxiang Zhang and Cheng-Lin Liu}
}

@misc{li2024comprehensivesurveyoraclecharacter,
      title={A comprehensive survey of oracle character recognition: challenges, benchmarks, and beyond}, 
      author={Jing Li and Xueke Chi and Qiufeng Wang and Dahan Wang and Kaizhu Huang and Yongge Liu and Cheng-lin Liu},
      year={2024},
      eprint={2411.11354},
      archivePrefix={arXiv},
      primaryClass={cs.CV},
      url={https://arxiv.org/abs/2411.11354}, 
}

@article{bai2025qwen3,
  title={Qwen3-vl technical report},
  author={Bai, Shuai and Cai, Yuxuan and Chen, Ruizhe and Chen, Keqin and Chen, Xionghui and Cheng, Zesen and Deng, Lianghao and Ding, Wei and Gao, Chang and Ge, Chunjiang and others},
  journal={arXiv preprint arXiv:2511.21631},
  year={2025}
}

@misc{wang2024opendatasetoraclebone,
      title={An open dataset for oracle bone script recognition and decipherment}, 
      author={Pengjie Wang and Kaile Zhang and Xinyu Wang and Shengwei Han and Yongge Liu and Jinpeng Wan and Haisu Guan and Zhebin Kuang and Lianwen Jin and Xiang Bai and Yuliang Liu},
      year={2024},
      eprint={2401.15365},
      archivePrefix={arXiv},
      primaryClass={cs.CV},
      url={https://arxiv.org/abs/2401.15365}, 
}

@article{Wang2026AttGraphDC,
  title={AttGraph disentangling confusable ancient Chinese characters via component-correlation synergy},
  author={Kaili Wang and Tianquan Wu and Yuanlin Shi and Chen Chen},
  journal={npj Heritage Science},
  year={2026},
  volume={14},
  url={https://api.semanticscholar.org/CorpusID:284553809}
}

@article{wang2025multi,
  title={Multi-modal ancient scripts recognition via deep learning with data homogenization and augmentation},
  author={Wang, Nan and Wang, Weichen and Li, Bang and Zhang, Han and Jiao, Qingju and Liu, Chaofan},
  journal={npj Heritage Science},
  volume={13},
  number={1},
  pages={522},
  year={2025},
  publisher={Springer International Publishing Cham}
}

@misc{wu2019memoryreplayganslearning,
      title={Memory Replay GANs: learning to generate images from new categories without forgetting}, 
      author={Chenshen Wu and Luis Herranz and Xialei Liu and Yaxing Wang and Joost van de Weijer and Bogdan Raducanu},
      year={2019},
      eprint={1809.02058},
      archivePrefix={arXiv},
      primaryClass={cs.CV},
      url={https://arxiv.org/abs/1809.02058}, 
}

@misc{chaudhry2019tinyepisodicmemoriescontinual,
      title={On Tiny Episodic Memories in Continual Learning}, 
      author={Arslan Chaudhry and Marcus Rohrbach and Mohamed Elhoseiny and Thalaiyasingam Ajanthan and Puneet K. Dokania and Philip H. S. Torr and Marc'Aurelio Ranzato},
      year={2019},
      eprint={1902.10486},
      archivePrefix={arXiv},
      primaryClass={cs.LG},
      url={https://arxiv.org/abs/1902.10486}, 
}

@article{Kirkpatrick_2017,
   title={Overcoming catastrophic forgetting in neural networks},
   volume={114},
   ISSN={1091-6490},
   url={http://dx.doi.org/10.1073/pnas.1611835114},
   DOI={10.1073/pnas.1611835114},
   number={13},
   journal={Proceedings of the National Academy of Sciences},
   publisher={Proceedings of the National Academy of Sciences},
   author={Kirkpatrick, James and Pascanu, Razvan and Rabinowitz, Neil and Veness, Joel and Desjardins, Guillaume and Rusu, Andrei A. and Milan, Kieran and Quan, John and Ramalho, Tiago and Grabska-Barwinska, Agnieszka and Hassabis, Demis and Clopath, Claudia and Kumaran, Dharshan and Hadsell, Raia},
   year={2017},
   month=mar, pages={3521–3526} }

@misc{li2017learningforgetting,
      title={Learning without Forgetting}, 
      author={Zhizhong Li and Derek Hoiem},
      year={2017},
      eprint={1606.09282},
      archivePrefix={arXiv},
      primaryClass={cs.CV},
      url={https://arxiv.org/abs/1606.09282}, 
}

@misc{rebuffi2017icarlincrementalclassifierrepresentation,
      title={iCaRL: Incremental Classifier and Representation Learning}, 
      author={Sylvestre-Alvise Rebuffi and Alexander Kolesnikov and Georg Sperl and Christoph H. Lampert},
      year={2017},
      eprint={1611.07725},
      archivePrefix={arXiv},
      primaryClass={cs.CV},
      url={https://arxiv.org/abs/1611.07725}, 
}

@misc{yu2020semanticdriftcompensationclassincremental,
      title={Semantic Drift Compensation for Class-Incremental Learning}, 
      author={Lu Yu and Bartłomiej Twardowski and Xialei Liu and Luis Herranz and Kai Wang and Yongmei Cheng and Shangling Jui and Joost van de Weijer},
      year={2020},
      eprint={2004.00440},
      archivePrefix={arXiv},
      primaryClass={cs.CV},
      url={https://arxiv.org/abs/2004.00440}, 
}

@article{mejias2025technical,
  title={Technical guidelines for digitizing cultural heritage materials [Gu{\'\i}a t{\'e}cnica para la digitalizaci{\'o}n de materiales de herencia cultural},
  author={Mej{\'\i}as, Natalia Hern{\'a}ndez},
  journal={Acceso. Revista Puertorrique{\~n}a de Bibliotecolog{\'\i}a y Documentaci{\'o}n},
  year={2025}
}

@article{qwen3vlembedding,
  title={Qwen3-VL-Embedding and Qwen3-VL-Reranker: A Unified Framework for State-of-the-Art Multimodal Retrieval and Ranking},
  author={Li, Mingxin and Zhang, Yanzhao and Long, Dingkun and Chen, Keqin and Song, Sibo and Bai, Shuai and Yang, Zhibo and Xie, Pengjun and Yang, An and Liu, Dayiheng and Zhou, Jingren and Lin, Junyang},
  journal={arXiv},
  year={2026}
}

@inproceedings{he2016deep,
  title={Deep residual learning for image recognition},
  author={He, Kaiming and Zhang, Xiangyu and Ren, Shaoqing and Sun, Jian},
  booktitle={Proceedings of the IEEE conference on computer vision and pattern recognition},
  pages={770--778},
  year={2016}
}

@inproceedings{radford2021learning,
  title={Learning transferable visual models from natural language supervision},
  author={Radford, Alec and Kim, Jong Wook and Hallacy, Chris and Ramesh, Aditya and Goh, Gabriel and Agarwal, Sandhini and Sastry, Girish and Askell, Amanda and Mishkin, Pamela and Clark, Jack and others},
  booktitle={International conference on machine learning},
  pages={8748--8763},
  year={2021},
  organization={PmLR}
}

@inproceedings{jia2021scaling,
  title={Scaling up visual and vision-language representation learning with noisy text supervision},
  author={Jia, Chao and Yang, Yinfei and Xia, Ye and Chen, Yi-Ting and Parekh, Zarana and Pham, Hieu and Le, Quoc and Sung, Yun-Hsuan and Li, Zhen and Duerig, Tom},
  booktitle={International conference on machine learning},
  pages={4904--4916},
  year={2021},
  organization={PMLR}
}

@article{garg2023tic,
  title={Tic-clip: Continual training of clip models},
  author={Garg, Saurabh and Farajtabar, Mehrdad and Pouransari, Hadi and Vemulapalli, Raviteja and Mehta, Sachin and Tuzel, Oncel and Shankar, Vaishaal and Faghri, Fartash},
  journal={arXiv preprint arXiv:2310.16226},
  year={2023}
}

@inproceedings{zhu2023ctp,
  title={Ctp: Towards vision-language continual pretraining via compatible momentum contrast and topology preservation},
  author={Zhu, Hongguang and Wei, Yunchao and Liang, Xiaodan and Zhang, Chunjie and Zhao, Yao},
  booktitle={Proceedings of the IEEE/CVF International Conference on Computer Vision},
  pages={22257--22267},
  year={2023}
}

@article{goswami2025query,
  title={Query drift compensation: Enabling compatibility in continual learning of retrieval embedding models},
  author={Goswami, Dipam and Wang, Liying and Twardowski, Bart{\'L} and van de Weijer, Joost and others},
  journal={arXiv preprint arXiv:2506.00037},
  year={2025}
}

@misc{buzzega2020darkexperiencegeneralcontinual,
      title={Dark Experience for General Continual Learning: a Strong, Simple Baseline}, 
      author={Pietro Buzzega and Matteo Boschini and Angelo Porrello and Davide Abati and Simone Calderara},
      year={2020},
      eprint={2004.07211},
      archivePrefix={arXiv},
      primaryClass={stat.ML},
      url={https://arxiv.org/abs/2004.07211}, 
}

@inproceedings{gordin2024cured,
  title={CuReD: Deep learning optical character recognition for Cuneiform text editions and legacy materials},
  author={Gordin, Shai and Alper, Morris and Romach, Avital and Santos, Luis Saenz and Yochai, Naama and Lalazar, Roey},
  booktitle={Proceedings of the 1st Workshop on Machine Learning for Ancient Languages (ML4AL 2024)},
  pages={130--140},
  year={2024}
}

@article{fuentes2025recognition,
  title={Recognition of Egyptian hieroglyphic texts through focused generic segmentation and cross-validation voting},
  author={Fuentes-Ferrer, Ra{\'u}l and Duque-Domingo, Jaime and Herrera, Pedro Javier},
  journal={Applied Soft Computing},
  volume={171},
  pages={112793},
  year={2025},
  publisher={Elsevier}
}

@inproceedings{swindall2021exploring,
  title={Exploring learning approaches for ancient Greek character recognition with citizen science data},
  author={Swindall, Matthew I and Croisdale, Gregory and Hunter, Chase C and Keener, Ben and Williams, Alex C and Brusuelas, James H and Krevans, Nita and Sellew, Melissa and Fortson, Lucy and Wallin, John F},
  booktitle={2021 IEEE 17th International conference on eScience (eScience)},
  pages={128--137},
  year={2021},
  organization={IEEE}
}

@inproceedings{wang2024human,
  title={Human-llm collaborative annotation through effective verification of llm labels},
  author={Wang, Xinru and Kim, Hannah and Rahman, Sajjadur and Mitra, Kushan and Miao, Zhengjie},
  booktitle={Proceedings of the 2024 CHI conference on human factors in computing systems},
  pages={1--21},
  year={2024}
}

@article{wang2024open,
  title={An open dataset for oracle bone character recognition and decipherment},
  author={Wang, Pengjie and Zhang, Kaile and Wang, Xinyu and Han, Shengwei and Liu, Yongge and Wan, Jinpeng and Guan, Haisu and Kuang, Zhebin and Jin, Lianwen and Bai, Xiang and others},
  journal={Scientific Data},
  volume={11},
  number={1},
  pages={976},
  year={2024},
  publisher={Nature Publishing Group UK London}
}

@article{wang2022oracle,
  title={Oracle-mnist: a dataset of oracle characters for benchmarking machine learning algorithms},
  author={Wang, Mei and Deng, Weihong},
  journal={arXiv preprint arXiv:2205.09442},
  year={2022}
}

@inproceedings{huang2019obc306,
  title={Obc306: A large-scale oracle bone character recognition dataset},
  author={Huang, Shuangping and Wang, Haobin and Liu, Yongge and Shi, Xiaosong and Jin, Lianwen},
  booktitle={2019 International Conference on Document Analysis and Recognition (ICDAR)},
  pages={681--688},
  year={2019},
  organization={IEEE}
}

@article{wang2023gan,
  title={A gan-based denoising method for chinese stele and rubbing calligraphic image},
  author={Wang, Xuanhong and Wu, Kun and Zhang, Ying and Xiao, Yun and Xu, Pengfei},
  journal={The visual computer},
  volume={39},
  number={4},
  pages={1351--1362},
  year={2023},
  publisher={Springer}
}
